%% file: main_full.tex
\documentclass[letterpaper]{article} \usepackage{aaai24}  \usepackage{times}  \usepackage{helvet}  \usepackage{courier}  \usepackage[hyphens]{url}  \usepackage{graphicx}    \usepackage{natbib}  \usepackage{caption}     \usepackage{algorithm}
\usepackage{algpseudocode}
\usepackage{adjustbox}
\usepackage{newfloat}
\usepackage{listings}
\DeclareCaptionStyle{ruled}{labelfont=normalfont,labelsep=colon,strut=off} 
\floatstyle{ruled}
\newfloat{listing}{tb}{lst}{}
\floatname{listing}{Listing}
\usepackage[utf8]{inputenc} \usepackage[T1]{fontenc}    \usepackage{url}            \usepackage{booktabs}       \usepackage{nicefrac}       \usepackage{microtype}      \usepackage{xcolor}         \usepackage[inline]{enumitem}       \usepackage{xspace}
\usepackage{placeins}
\usepackage{lipsum}
\usepackage{bm}
\usepackage{float}
\usepackage{graphicx}
\usepackage{multirow}
\usepackage{pifont}
\usepackage{longtable}
\usepackage{subcaption}
\usepackage{amsmath}
\usepackage{amsfonts}
\usepackage{chngcntr}

\usepackage{xfrac}

\title{Colored Noise in PPO: Improved Exploration and Performance through Correlated Action Sampling}

\nocopyright

\author {
Jakob Hollenstein\textsuperscript{\rm 1},
    Georg Martius\textsuperscript{\rm 3,4},
    Justus Piater\textsuperscript{\rm 1,2}
}
\affiliations {
\textsuperscript{\rm 1}Department of Computer Science, University of Innsbruck, Austria\\
    \textsuperscript{\rm 2}Digital Science Center, University of Innsbruck, Austria\\
    \textsuperscript{\rm 3}Max Planck Institute for Intelligent Systems, Tübingen, Germany\\
    \textsuperscript{\rm 4}Department of Computer Science, University of Tübingen, Germany\\
\{jakob.hollenstein,justus.piater\}@uibk.ac.at\\
    georg.martius@uni-tuebingen.de\\
}

\newcommand{\ANALYSISPATH}{analysis/aaai}
\newcommand{\eclearpage}{\newpage}
\newcommand{\aaaiversion}{}

\usepackage{xr-hyper}
\makeatletter
\newcommand*{\addFileDependency}[1]{\typeout{(#1)}
  \@addtofilelist{#1}
  \IfFileExists{#1}{}{\typeout{No file #1.}}
}
\makeatother

\newcommand{\cmark}{\ding{51}}\newcommand{\xmark}{\ding{55}}\newcommand{\nenv}{N_\textrm{env}}\newcommand{\nsteps}{N_\textrm{steps}}\newcommand{\nepochs}{N_\textrm{epochs}}

\newcommand{\refalg}[1]{{Algorithm \ref{#1}}}
\newcommand{\refeq}[1]{{(\ref{#1})}}
\newcommand{\refsec}[1]{{Sec.~\ref{#1}}}
\newcommand{\reffig}[1]{{Figure~\ref{#1}}}
\newcommand{\refsubfig}[1]{{(\subref{#1})}}
\newcommand{\reftbl}[1]{{Table~\ref{#1}}}

\DeclareMathOperator*{\E}{\mathbb{E}}

\newcommand{\changed}[1] {{\color{blue} #1}}
 \renewcommand{\changed}[1] {{#1}}

\makeatletter
\begin{document}

\maketitle

\begin{abstract}
   Proximal Policy Optimization (PPO), a popular on-policy deep
 reinforcement learning method, employs a stochastic policy for
 exploration. In this paper, we propose a colored noise-based
 stochastic policy variant of PPO. Previous research highlighted the
 importance of temporal correlation in action noise for effective
 exploration in \emph{off-policy} reinforcement learning. Building on
 this, we investigate whether correlated noise can also enhance
 exploration in \emph{on-policy} methods like PPO. We discovered that
 correlated noise for action selection improves learning performance
 and outperforms the currently popular uncorrelated white noise
 approach in on-policy methods. Unlike off-policy learning, where pink
 noise was found to be highly effective, we found that a colored
 noise, intermediate between white and pink, performed best for
 on-policy learning in PPO. We examined the impact of varying the
 amount of data collected for each update by modifying the number of
 parallel simulation environments for data collection and observed
 that with a larger number of parallel environments, more strongly
 correlated noise is beneficial. Due to the significant impact and
 ease of implementation, we recommend switching to correlated noise as
 the default noise source in PPO.
 \end{abstract}

\newcommand\blfootnote[1]{\begingroup
  \renewcommand\thefootnote{}\footnote{#1}\addtocounter{footnote}{-1}\endgroup
}

\urldef{\urlPaper}\url{https://doi.org/10.1609/aaai.v38i11.29139}
\urldef{\urlCode}\url{https://github.com/jkbjh/cn-ppo-paper-code}
\blfootnote{This paper was originally published in: Proceedings of the
  AAAI Conference on Artificial Intelligence, 38. Available at:
  \mbox{\urlPaper}. This version contains the supplementary material; Code is available at: \mbox{\urlCode}.
}

\input{\ANALYSISPATH/better_default_count_commands.tex}

\section{Introduction}

Exploration plays a crucial role in deep reinforcement learning,
particularly in continuous action space applications like robotics,
where the number of states and actions is infinite. In the continuous
action space setting, exploration is typically achieved by introducing
variability around the mean action proposed by the policy. For
instance, deterministic off-policy algorithms such as TD3
\citep{fujimotoAddressingFunctionApproximation2018} and DDPG
\citep{lillicrapContinuousControlDeep2016} use additive action noise,
while stochastic off-policy algorithms such as SAC
\citep{haarnojaSoftActorCriticAlgorithms2019} and MPO
\citep{abdolmalekiMaximumPosterioriPolicy2018} employ a Gaussian
distributed policy to sample actions and induce variation. On-policy
algorithms such as TRPO \citep{schulmanTrustRegionPolicy2015} and PPO
\citep{schulmanProximalPolicyOptimization2017} follow a similar
approach by sampling variations from a Gaussian distribution.

While previous research \citep{hollensteinActionNoiseOffPolicy2022,
  eberhardPinkNoiseAll2023, raffinSmoothExplorationRobotic2021} has
  emphasized the importance of temporal correlation in action noise
  for exploration in \changed{\emph{off-policy}}
  algorithms, \changed{\emph{on-policy}} algorithms such as PPO do not
  incorporate correlated action noise and instead sample uncorrelated
  variations from a Gaussian distributed policy.

PPO, as an on-policy deep reinforcement learning algorithm, offers
several advantages over off-policy algorithms: by not relying on
replay buffers, on-policy algorithms exhibit fewer instabilities due
to distributional shift between the distribution of previously
collected data and the current policy-induced distribution. Moreover,
on-policy algorithms mitigate Q-function divergence issues since the
policy evaluation can leverage data collected under the current
policy. These benefits make on-policy methods particularly
advantageous when environment samples are inexpensive to
obtain. However, exploration in on-policy methods needs to be induced
by changes to the environment (exploration reward) or by changes to
the policy itself, as, by definition, on-policy methods learn from
data induced by the current policy. Consequently,
altering the exploration technique without changing the environment
implies a change in the policy distribution.

\paragraph{Contributions}
In this paper, we \changed{extend previous work on correlated
action-noise in \emph{off-policy} RL to \emph{on-policy} RL methods}:
we introduce and empirically evaluate a modification to PPO
incorporating temporally correlated colored noise into the stochastic
policy's distribution. Utilizing the re-parameterization trick, we
maintain a Gaussian distributed behavior while injecting correlated
noise instead of uncorrelated noise.
\changed{This introduces a new hyperparameter, the \emph{noise color}
$\beta$ parameterizing the correlated noise}.

Our experiments demonstrate that adopting colored noise enhances
performance in the majority of tested environments.
\changed{In the off-policy setting \citet{eberhardPinkNoiseAll2023} found a common
noise color, pink noise, to significantly outperform the previous
default of uncorrelated white noise. Surprisingly we}
found the optimal \changed{common} colored noise for improved
performance --- for the
\emph{on-policy} method PPO --- to lie \emph{between} white and pink
noise ($\beta=0.5$, see \refsec{sec:method}).
Furthermore, \changed{we investigated the effect of update dataset
size on the efficacy of the noise colors:} as the number of parallel
data collection environments increases\changed{, and thus the update
dataset size increases}, we observe a trend towards more correlated
noise\changed{. This provides a plausible explanation for the preference
of more correlated noise in the off-policy setting}. However, for the
benchmarks considered in this work, our results indicate that
utilizing about four parallel environments, resulting in $8192$
samples per update step (see \refsec{sec:nenv_vs_nstep} for the
relation of $\nsteps$ and $\nenv$), together with noise in between
white and pink noise, is the most efficient approach.

\subsection{Related Work}

Reinforcement learning is a promising technique for solving complex
control problems. An important milestone for the development of deep
reinforcement learning was the work of
\citet{mnihHumanlevelControlDeep2015} showing human level performance
of reinforcement learning on the Atari Games benchmark. In contrast to
the Atari Games benchmarks which has complicated observation spaces
but discrete action spaces, reinforcement learning for robotic
applications has to deal with continuous action spaces, for example in
the benchmarks provided by \citet{tassaDeepMindControlSuite2018}, where
the task is to control simulated robots. Because the action selection
process is more complicated in continuous spaces, policy search
methods \citep{williamsSimpleStatisticalGradientfollowing1992} are
favored. This includes off-policy methods such as DDPG
\citep{lillicrapContinuousControlDeep2016}, TD3
\citep{fujimotoAddressingFunctionApproximation2018}, MPO
\citep{abdolmalekiMaximumPosterioriPolicy2018} or SAC
\citep{haarnojaSoftActorCriticAlgorithms2019} which can learn from
arbitrary data and on-policy methods such as TRPO
\citep{schulmanTrustRegionPolicy2015} and PPO
\citep{schulmanProximalPolicyOptimization2017} which iteratively
improve the policy using data collected by the current iteration of
the policy. On-policy methods exhibit better convergence properties
and behave more stably but require larger amounts of training data.

Exploration, the problem of discovering better action sequences, is of
pivotal importance in RL and is tackled in many
ways. \citet{aminSurveyExplorationMethods2021},
\citet{ladoszExplorationDeepReinforcement2022} and
\citet{yangExplorationDeepReinforcement2022} provide recent surveys of
the exploration research landscape in deep reinforcement learning.  In
this work, we turn to the simple yet effective method which forms the
backbone of most RL algorithms: undirected noisy exploration.

In off-policy methods, exploration is often achieved by perturbing the
action selection process, for example by perturbing the parameters of
the policy \citep{plappertParameterSpaceNoise2018} or by additively
perturbing the action, i.e., by adding \emph{action noise}.  This
action perturbation can be done, for example, using temporally
uncorrelated Gaussian noise, or temporally correlated noise generated
by an Ornstein-Uhlenbeck process
\citep{uhlenbeckTheoryBrownianMotion1930}. Previous research has shown
that \changed{increasing the } temporal correlation of actions,
i.e.,\ \changed{switching from} Gaussian noise
\changed{to} Ornstein-Uhlenbeck noise\changed{, tends to increase state space
coverage, but an increase in state space coverage can be beneficial or
harmful for policy learning; this depends on the task environment
and its dynamics}
\citep{hollensteinHowDoesType2021,hollensteinActionNoiseOffPolicy2022}.
Another method for temporally correlated perturbations was introduced
by \citet{ruckstiessStateDependentExplorationPolicy2008,raffinSmoothExplorationRobotic2021} where the ``action
noise'' is generated deterministically by a function only dependent on
the state, but the function parameters are randomly changed after a
number of steps. \changed{This approach is extended
by \citet{chiappaLatentExplorationReinforcement2023} to exploit the
hidden layers of the policy, to induce correlation between the action
dimensions.}

While different environments respond differently to the temporal
correlation of noise, \mbox{\citet{eberhardPinkNoiseAll2023}} found
\emph{pink noise} to act as a kind of middle ground between
uncorrelated white Gaussian noise, and correlated Brownian-motion-like
noise, e.g.,\ red noise and Ornstein-Uhlenbeck noise. This middle
ground, while not always the perfect choice, was found to be a much
better default choice.

\citet{petrazziniProximalPolicyOptimization2021} propose to improve
exploration by changing the policy distribution of PPO from a Gaussian
distribution to a Beta distribution. Similarly, we propose to change
the policy distribution of PPO, but instead of a Beta distribution we
propose to keep the Gaussian distribution and bias the sampling to be
temporally correlated using a colored-noise process.

Algorithmic details and hyperparameter choices are known to be
important for PPO and were empirically analyzed by
\citet{andrychowiczWhatMattersOnPolicy2021} and
\citet{engstromImplementationMattersDeep2020}. Similarly, we performed
a comprehensive empirical evaluation of the impact of different noise
correlation settings and the setting of parallel data collection. From
this evaluation, we recommend switching the default noise process to a
correlated noise process with $\beta=0.5$.

\section{Method\label{sec:method}}
\begin{figure}[b]
  \centering
  \includegraphics[width=\linewidth]{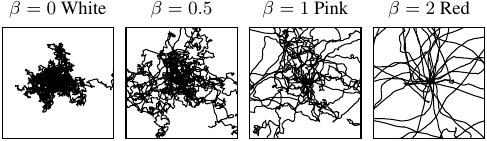}
  \caption{Two-dimensional random walks caused by colored noise of
    different $\beta$. Lower $\beta$ values cause more energy in high
    frequency parts of the power spectral density, causing the random
    walk to change direction more frequently and thus causing more
    local and less global exploration. Higher values of $\beta$ result
    in more energy in the lower frequencies of the power spectral
    density. This translates to random walks that change direction
    less frequently, and thus explore more globally. }
  \label{fig:noise_properties_random_walks}
\end{figure}

\subsection{Colored Noise}
Colored noise is a class of noise that exhibits temporal correlation
and is characterized by a change of $\frac{1}{f^{\beta}}$ in its power
spectral density (PSD) components, where $f$ denotes the frequency and
$\beta$ determines the ``color'' of the noise. The power spectral
density (PSD) of a sequence consists of the squared magnitude of the
frequency components of the sequence's Fourier transform.  Given
sequences of noise samples $\tau^{(i)}_\varepsilon =
\{\varepsilon_1^{(i)}, \ldots \varepsilon_T^{(i)}\}$, and the power
spectral density of each sequence
$|\mathcal{F}(\tau^{(i)}_\varepsilon)|^2$, the \emph{expected PSD} is
calculated by averaging over each trajectory's PSD:
$\E_i[|\mathcal{F}(\tau_\varepsilon^{(i)})|^2]$.

Uncorrelated noise is called \emph{white noise} and has a constant,
flat line, expected power spectral density ($\beta=0$).  The expected
PSD of \emph{pink noise} decreases with $\sfrac{1}{f}$ ($\beta=1$) and
for \emph{red noise} the decrease happens at a rate of
$\sfrac{1}{f^2}$ ($\beta=2$).

The effect of sampling actions from colored noise processes of
different colors can be exemplified by a velocity controlled robot
that can take steps of various sizes in the $x$ and $y$
directions. The paths taken by the robot, when the steps are
controlled by noise, depend on the correlation of the noise. The
position ($x_t, y_t$) of the robot is the result of integrating over
the noise sequence. Examples of such integrated sequences,
\emph{random walks}, for colored noises of different $\beta$
coefficients are shown in
\reffig{fig:noise_properties_random_walks}. Brownian motion, for
example, exhibits red noise ($\beta=2$) characteristics and is itself
the result of integrating white noise ($\beta=0$) over time.

Colored noise can be generated in the time domain by drawing white
noise samples and applying an autoregressive filter or directly in the
frequency domain by sampling the frequency components accordingly. In
our work we use the latter approach, by building on an implementation
by \citet{patzeltFelixpatzeltColorednoise2019} and adapted by
\citet{eberhardPinkNoiseAll2023} following an algorithm by
\citet{timmerGeneratingPowerLaw1995}. We generate colored noise
sequences of length $\tau$: $\tau_\varepsilon = \{\varepsilon_1,
\ldots, \varepsilon_\tau\}$. Sampling action noise is then implemented
as consuming the items from this sequence. \refsec{sec:cngen} for
further details.

\subsection{PPO}
PPO is a popular on-policy reinforcement learning algorithm that utilizes
a policy gradient approach to optimize the policy. To avoid learning
instabilities, PPO aims to prevent large policy updates. As an on-policy
algorithm, PPO collects trajectories by interacting with the
environment using the current policy. Advantages are then computed for
these data and the policy is updated. For the next policy improvement,
new data is collected. The number of data points collected for each
update is a hyperparameter. 

\begin{samepage}
The action selection process uses a policy network that outputs the
mean ($\mu_t$) and standard deviation~($\sigma_t$) of a Gaussian
distribution. An action is then sampled from this distribution:
\begin{equation}
  a_t \sim \mathcal{N}(\mu_t, \sigma_t).
\end{equation}
Using the re-parameterization trick, this is re-written:
\begin{equation}
  a_t = \mu_t + \varepsilon_t \cdot \sigma_t; ~~~ \varepsilon_t \sim \mathcal{N}(0, 1).
  \label{eq:reparameterization}
\end{equation}
Instead of sampling $\varepsilon_t$ from a white noise Gaussian
process, the default in PPO, we propose to use a colored noise
Gaussian process, similar to the approach of
\citet{eberhardPinkNoiseAll2023} for \emph{off-policy} methods.
\end{samepage}

The generated noise samples $\varepsilon_t^{(i)}$ still show Gaussian
distributions at each time step
(\reffig{subfig:noise-has-gaussian-limit}). Because we modify the
$\varepsilon$ in $\mu + \sigma \cdot \varepsilon$ and $\varepsilon_t$
remains Gaussian, the data collection, viewed at each individual step,
remains asymptotically on-policy. However, while the marginal stays
Gaussian, the two-step correlation of $\varepsilon_t$
vs. $\varepsilon_{t+1}$ changes with $\beta$
(\reffig{subfig:gaussian_correlated}). In this paper, we empirically
evaluate the impact of this correlation due to the use of colored
noise.

\section{Experiments}

\begin{figure}
  \includegraphics[width=\columnwidth]{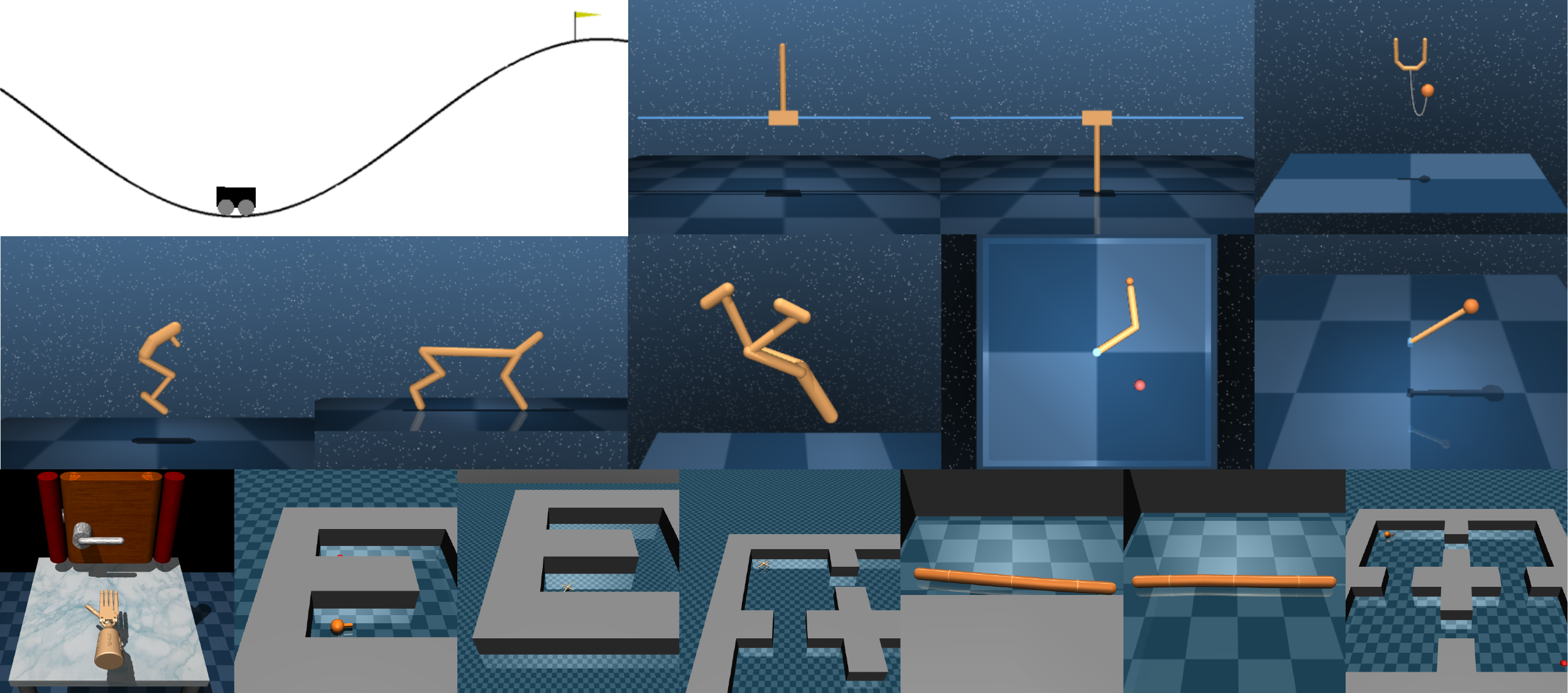}
  \caption{Benchmark environments:
    (top) Mountain Car, Cartpole
    Balance, Cartpole Swingup, Ball in Cup (Catch), (middle) Hopper
    Hop, Cheetah Run, Walker Run, Reacher Hard, Pendulum Swingup, (bottom) Door,
    UMaze Point, UMaze Ant, 4 Rooms Point, UMaze Swimmer, 4 Rooms Swimmer, 4 Rooms Ant
    }
    \label{fig:environment_images}
\end{figure}

We perform an empirical evaluation of the impact of colored noise on
the performance of PPO. To this end we perform training runs, repeated
with $20$ independent seeds, on the benchmarks
\input{\ANALYSISPATH/used_env_names.tex}\unskip. We vary the noise
color $\beta \in \{$\input{\ANALYSISPATH/used_noise_colors.tex}$\}$
and test different numbers of parallel collection environments
$N_\textrm{envs} \in
\{\input{\ANALYSISPATH/used_num_envs.tex}\}$. This results in a total
of $\input{\ANALYSISPATH/total_number_of_experiments.tex}$
experiments. See \refsec{sec:experiment_details} for further details.

\subsection{Evaluation Details}
\label{sec:evaluation_details}
We train each agent for a total of $2\,048\,000$ time steps and
evaluate every $10240$ steps, resulting in $200$ evaluation points. At
each evaluation point, $50$ evaluation episodes are performed. We average
the mean returns collected at each evaluation point, forming an estimate
that captures the area under the learning curve, and refer to this as the
\emph{performance}. Evaluation results are grouped by environment and
standardized to zero mean and unit variance to control for the impact of
the environments. We combine these standardized results by averaging across
different environments and seeds.

\FloatBarrier
\subsubsection{Does Colored-Noise Affect the Performance of PPO?}

Adding action-noise to the action selection process creates a
divergence between the action distribution of the policy and the
actual action distribution, i.e., it induces a difference between the
distribution of the data the undisturbed policy would collect,
compared to the policy disturbed by additive action noise.  Similarly,
modifying the action selection process, by changing $\varepsilon$ in
\refeq{eq:reparameterization}, can potentially break the learning
process, as on-policy algorithms operate under the assumption that the
collected data match the distribution induced by the policy.

\begin{figure}
  \includegraphics[width=\linewidth]{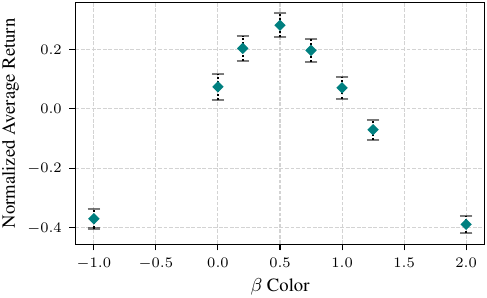}
  \caption{Performance averaged across Environments: Correlated noise $\beta = 0.5$
    significantly outperforms the default white noise ($\beta=0$, \refsec{sec:beta0vanilla}) used by PPO. The     bars indicate the $95\%$ bootstrapped confidence
    intervals.}
  \label{fig:average_returns_vs_color}
\end{figure}

In this experiment, we compare how the learning performance (area
under the learning curve, averaged across environments,
\refsec{sec:evaluation_details}) of PPO reacts to a modified action
selection process: to the use of correlated noise ($\beta\neq 0$) as
compared to the standard case of using uncorrelated noise ($\beta =
0$). \reffig{fig:average_returns_vs_color} shows the performance
averaged across the different environments and across the different
numbers of parallel data collection environments. Results for each
environment are standardized to control for differences in reward
scale. The mean performance, and $95\%$ confidence intervals of the
mean, are depicted. The confidence intervals are estimated using
(bias-corrected and accelerated) bootstrapping. The results indicate a
significant effect of the correlation of the noise. This is in line
with previous findings highlighting an impact of action-noise
correlation on learned performance
\citep{hollensteinActionNoiseOffPolicy2022,eberhardPinkNoiseAll2023}.
The results depicted in \reffig{fig:average_returns_vs_color} show an
overall preference for colored noise at $\beta=0.5$, which lies
between white noise ($\beta=0$) and pink noise ($\beta=1$). This
contrasts with the results in off-policy learning, where $\beta=1$ was
found to be superior across many environments
\citep{eberhardPinkNoiseAll2023}.  On-policy methods assume matching
state-visitation frequencies between the collected data and what the
current policy would induce.  More correlated noise induces larger
state space coverage \citep{hollensteinActionNoiseOffPolicy2022} and
thus a larger deviation from the states the deterministic-mean-action
policy would visit.  Thus, a potential reason for this difference in
noise color preference is that increasing the state-space coverage, by
increasing the noise correlation, increases the distributional shift
between the data and policy-induced marginal state-visitation
frequencies.

In summary, we found colored noise to have a
significant impact on learned policy performance averaged across
environments, with $\beta=0.5$ performing best.

\subsubsection{Is $\beta=0.5$ a Better Default for PPO?}

The default stochastic policy for PPO relies on uncorrelated noise
sampled from a Gaussian distribution. The results shown in
\reffig{fig:average_returns_vs_color} indicate that across
environments $\beta=0.5$ performs best. Previous research found an
environment specific response to action noise correlation for purely
noisy policies \citep{hollensteinHowDoesType2021}, as well as learning
performance in off-policy methods
\citep{hollensteinActionNoiseOffPolicy2022,eberhardPinkNoiseAll2023}.
In accordance with these earlier results, we found the best $\beta^*$
to vary depending on the
environment. \reftbl{tbl:best-noise-color-per-environment} lists the
best performing $\beta^*$ for each environment. In addition, we
perform Welch t-tests to compare the performance of $\beta^*$ to
each $\beta \in \{0.5, 0.0, 1.0\}$. This shows that $\beta=0.5$
performs comparable to $\beta^*$ in $\isfavorablebetaIntermediate/\betterdefaultnumexperiments$ environments. In contrast,
the current default $\beta=0$ is significantly outperformed by
$\beta^*$ in \isoutperformedbetaWhiteW environments and only performs comparable in $\isfavorablebetaWhite/\betterdefaultnumexperiments$
environments. The best default found for off-policy learning, pink
noise ($\beta=1.0$), is significantly outperformed in \isoutperformedbetaPinkW out of \betterdefaultnumexperimentsW
cases and performs comparable in $\isfavorablebetaPink/\betterdefaultnumexperiments$ environments.
\emph{We therefore recommend switching the default from temporally uncorrelated noise
$\beta=0$ to temporally correlated noise $\beta=0.5$.}

\begin{table}[t]
\centering
    {
    \footnotesize\setlength{\tabcolsep}{3pt}
    \input{\ANALYSISPATH/best-noise-color-per-environment.tex}}
\caption{Optimal noise color per environment.
  \cmark~vs.\ \xmark{} indicate whether each fixed $\beta
    \in \{0.5, 0.0, 1.0\}$ performs comparable to $\beta^*$ (\cmark)
    or is significantly (Welch t-test) outperformed by $\beta^*$
    (\xmark). The p-values are listed in $p_{0.5}, p_{0.0}$ and
    $p_{1.0}$. ~~ $\beta=0.5$ performs favorable (\cmark $\isfavorablebetaIntermediate/\betterdefaultnumexperiments$) and
    improves over the current default of white noise $\beta=0$~(\cmark $\isfavorablebetaWhite/\betterdefaultnumexperiments$)}
  \label{tbl:best-noise-color-per-environment}
\end{table}

\subsubsection{How Does the Number of Parallel Collection Environments Affect the Performance?\label{sec:effect-of-parallel-envs}}
\begin{figure}[!t]
\includegraphics[width=\linewidth]{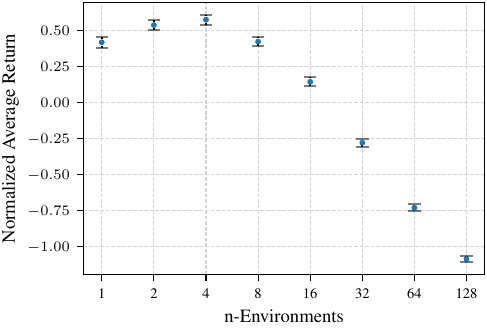}
\caption{Performance averaged across environments and noise colors: the number of
    parallel data collection environments has a significant
    impact on the performance. Bootstrapped $95\%$ confidence
      intervals for the mean are shown. With $\nenv=4$ achieving the
      highest performance, though not significantly outperforming $\nenv=2$}
  \label{fig:averaged_returns_vs_num_envs}
\end{figure}

On-policy methods are less efficient with respect to environment
interaction samples compared to off-policy methods. This makes
on-policy methods particularly interesting when environment samples
are cheap to collect and can be collected in parallel. We collected
environment samples with different numbers of parallel environments
$N_\textrm{envs} \in \{\input{\ANALYSISPATH/used_num_envs.tex}\}$ and kept
the number of total environment interactions (total time steps)
constant across experiments. With each of the parallel environments,
$2048$ samples are collected as the dataset for each update
cycle. Thus, depending on the number of environments, more (or less)
samples are used for each update. Because the total number of samples
is limited (\refsec{sec:evaluation_details}) this implies fewer (or
more) updates in total. \reffig{fig:averaged_returns_vs_num_envs}
indicates that a larger number of parallel environments negatively
affects performance. This is in line with findings by
\citet{andrychowiczWhatMattersOnPolicy2021}, who also found that the
most beneficial number of parallel environments is dependent on the
environment type. In our study, we find that about four parallel
environments are preferable. This translates to a preference for about
$2048 \cdot 4 = 8192$ samples in each update if the episode lengths
are small enough to fit the $2048$ step limit. In summary, we found
8192 samples per update, collected by four environments in parallel,
to perform most efficiently for the investigated class of tasks.

\subsubsection{How Does the Number of Parallel Collection Environments and Noise Color Interact?}
The previous section indicated that the learning performance is
impacted by the number of parallel collection environments and hence
the size of the dataset used in each update cycle. We found a
significant difference between the noise preference in off-policy
methods ($\beta=1.0$) and PPO as an on-policy method
($\beta=0.5$). Since off-policy methods showed a preference for a
larger $\beta$ ($\beta=1$) and off-policy methods employ a replay
buffer, and thus have access to more samples in each update, this begs
the question: does the number of samples used in each PPO update show
any interaction with the preferred noise type $\beta$?

\begin{figure}[!t]
  \begin{subfigure}{.48\linewidth}
    \includegraphics[width=\linewidth]{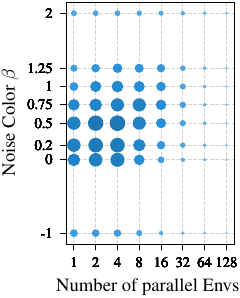}
    \subcaption{Marker size represents averaged performance (scaled to the sixth power to highlight differences)}
    \label{subfig:marker_size_average_performance}
  \end{subfigure}
  ~~
  \begin{subfigure}{.48\linewidth}
    \includegraphics[width=\linewidth]{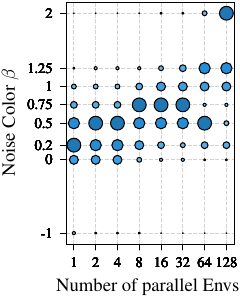}
    \subcaption{Marker size represents rank of averaged performance
      within the same number of environments}
    \label{subfig:marker_size_rank_performance}
  \end{subfigure}
  \caption{Preferred noise color depends on number of environments:
    Average performance across environment, impact of noise color
    $\beta$ combined with $\textrm{n-envs}$ number of parallel
    environments: \refsubfig{subfig:marker_size_average_performance} A trend
    is visible: the averaged performance is larger for larger $\beta$
    when more collection environments are used, but the decline due to
    the number of environments outweighs this
    trend. \refsubfig{subfig:marker_size_rank_performance} Ranks of average
    performance are indicated by circle size, ranks are calculated
    across noise-colors but within the same number of
    environments. The positive trend between number of environments
    and larger $\beta$ is clearly visible.  }
  \label{fig:noise-color-vs-n-envs-rank}
\end{figure}

\reffig{subfig:marker_size_average_performance} shows the average
performance across environments depending on the number of collection
environments (x) and noise color (y).
A positive association between larger $\beta$ (more correlated noise)
and larger number of environments $\nenv$ becomes visible. This trend
is more discernible when the performances for each noise color are
compared separately for each number of parallel collection
environments. \reffig{subfig:marker_size_rank_performance} shows the
rank of the average performance, separately ranked for each color:
a larger number of environments leads to favorable performance of more
correlated noise, indicating that with larger samples (i.e.,\ larger
$\nenv$), more exploration is beneficial. However, in combination with
the finding from \refsec{sec:effect-of-parallel-envs}, we find the
best performing configuration as $\beta=0.5, N_{\textrm{envs}}=4$.

\begin{figure}[!t]
  \begin{subfigure}{0.49\linewidth}
    \includegraphics[width=\linewidth]{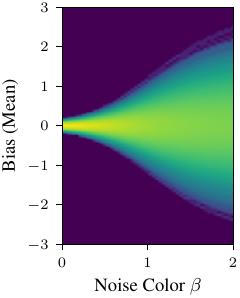}
    \caption{Distribution of the bias in $\varepsilon_t$, calculated
      by the mean across sequences of perturbations
      $\{\varepsilon_0, \ldots, \varepsilon_T\}$, as a function of
    the noise color $\beta$.}
    \label{subfig:bias_distribution}
  \end{subfigure}
  \hfill
    \begin{subfigure}{0.49\linewidth}
      \includegraphics[width=\linewidth]{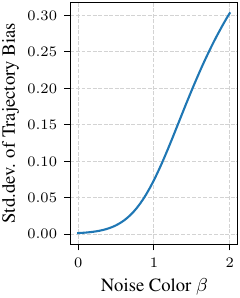}
      \caption{Standard deviation of the bias in $\{\varepsilon_0,
        \ldots, \varepsilon_T\}$, calculated as the standard
        deviations over the biases in
        \refsubfig{subfig:bias_distribution} as a function of noise
        color $\beta$.}
      \label{subfig:bias_stddeviation}
  \end{subfigure}
   \caption{Increasing the noise color $\beta$, increases the
     correlation in the perturbations and the spread of the bias of
     each noise sequence.}
\end{figure}

\subsubsection{Why Do Larger $\beta$ Work Better for Larger $\nenv$?}

\begin{figure}
\includegraphics[width=1.\linewidth]{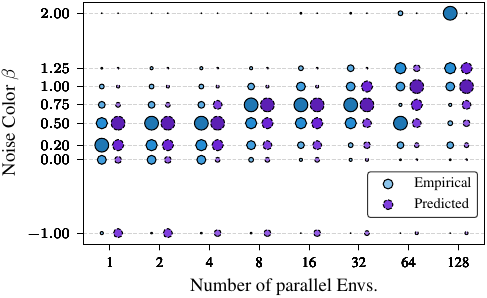}
  \caption{Performance rank, viewed separately for each number of
    parallel environments. Results from our empirical analysis are
    compared to a projected best noise color. Larger numbers of
    parallel environments decrease the variance of the bias, allowing
    for larger $\beta$.  }
  \label{fig:empirical_project_bias_beta}
\end{figure}

In the previous section we showed that, grouping for each $\nenv$ and
comparing the different $\beta$ within each group, there is a tendency
for better performance in larger $\beta$ values when $\nenv$ was
increased.

Larger $\beta$ values lead to more correlated perturbation sequences
$\{\varepsilon_1, \ldots \varepsilon_T\}$ and, since many environments
feature integrative dynamics, larger state space coverage
\citep{eberhardPinkNoiseAll2023,hollensteinActionNoiseOffPolicy2022}. While
this is beneficial for acquiring new information, it also implies that
the samples are more spread out and that the density around the mean
actions proposed by the policy is smaller. Similarly, a systematic
non-zero bias would prevent the policy from collecting on-policy data and
would lead to learning instability.

We measured the bias, as the mean across each sequence of
$\{\varepsilon_1, \ldots, \varepsilon_T\}$, for different $\beta$ and
found that, while there is no systematic bias ($=0$) for all $\beta$,
the spread of the bias increases with $\beta$
(\reffig{subfig:bias_distribution}). The resulting standard deviation
of the bias as a function of $\beta$ is shown in
\reffig{subfig:bias_stddeviation}.

When $\nenv$ is increased, more sequences of $\{\varepsilon_1, \ldots,
\varepsilon_T\}$ are pooled. This results in a decrease of the
variance of the bias of the collected sample $\sigma_N =
\frac{\sigma}{\sqrt{\nenv}}$. We can estimate a standard deviation
$\hat\sigma(\beta, \nenv)$ of the bias for a given $\beta$ and $\nenv$,
by combining $\sigma(\beta)$, the standard deviation of the bias as a
function of $\beta$ (\reffig{subfig:bias_stddeviation}), and the
reduction due to the sample size $\frac{1}{\sqrt{\nenv}}$.

We assume that the variance of the bias was optimal for the best
performing $\beta$ for each $\nenv$. We estimate this optimal variance
of the bias $\sigma^*$ as the average of $\hat\sigma(n, \beta)$ for
the best performing $\beta$~(\reftbl{tbl:best-beta-for-nenv}).
We calculate the difference between all combinations and the optimal
combination: ${E_{ij} = (\sigma^* - \hat\sigma(\beta_i, N_j))^2}$ and
rank these differences $E_{ij}$ separately for each $\nenv$
(\reffig{fig:empirical_project_bias_beta}). We find a trend
closely matching the performance ranking: larger variance due to
larger $\beta$ is compensated by larger number of environments (and
thus samples). The \emph{theoretical} best $\beta$ for each $\nenv$
(\reffig{fig:empirical_project_bias_beta}) closely follows our
\emph{empirical} observation
(\reffig{subfig:marker_size_rank_performance}).

This indicates, that the increased variance in the bias due to larger
$\beta$ is compensated by larger $\nenv$ and since generally more
exploration (i.e., larger $\beta$) would be favorable to collect more
diverse samples, the set of best trade-off configurations of $(\beta,
\nenv)$ shows a positive correlation between $\beta$ and $\nenv$.
\FloatBarrier

\section{Conclusion}

Previous work has found temporally correlated noise to be an effective
tool for noisy exploration in off-policy reinforcement learning. In
this work, we investigated whether temporally correlated noise can
also be applied to the on-policy method PPO. We propose to modify PPO
by employing the re-parameterization trick, to sample $\varepsilon$
from a Gaussian distributed, yet temporally correlated, noise
source. We performed a large empirical analysis
(in total $\input{\ANALYSISPATH/total_number_of_experiments.tex}$ experiments,
in \betterdefaultnumexperimentsW environments) and found
\begin{enumerate*}[label=(\roman*)]
    \item that colored noise indeed has a significant impact on the
      learning performance with colored noise ($\beta=0.5$)
      intermediate of white noise ($\beta=0$) and pink noise
      ($\beta=1$) to perform best, when averaging across different
      environments.
    \item We found that, while the best noise color $\beta$ depends on
      the environment, $\beta=0.5$ is favorable in
      \isfavorablebetaIntermediateW out of the
      \betterdefaultnumexperimentsW tested environments, and conclude
      that $\beta=0.5$ is a \emph{better default} noise color.
    \item We also investigated the effect of varying the update sample
      size by varying the number of parallel data collection
      environments and found $\nenv=4$ ($8192$ samples per update) to
      perform best when averaged across environments (and $\nenv \in
      \{2, 4\}$ significantly outperforming the other options).
    \item Interestingly, we found that with larger $\nenv$, and thus
      larger update sample size, the preference moves toward more
      correlated noise (i.e., larger $\beta$).
    \item We hypothesize that larger $\beta$ increase the uncertainty
      in the collected data: the variance of the effective bias over
      each action noise sequences increases with larger $\beta$ and
      larger sample sizes counteract the variance. We observe that the
      best performing $\beta$ as a function of the update size (by
      varying $\nenv$) follows the trend predicted by the change of
      variance in the bias as a function of $\beta$.
\end{enumerate*}

In summary, the results from our large empirical evaluation indicate
that $\beta=0.5$ is a better default choice for the noise in PPO. We
thus recommend switching the default noise source choice in PPO to
colored noise with $\beta=0.5$.

\section*{Acknowledgments}

We want to thank Onno Eberhard and Samuele Tosatto for helpful
feedback on earlier revisions of this text. Georg Martius is a member
of the Machine Learning Cluster of Excellence, EXC number 2064/1 –
Project number 390727645.

\bibliography{auto-generated-cn-ppo-processed}
 
\clearpage
\raggedbottom
\appendix
\section*{Supplementary Appendix}
\renewcommand{\thesection}{A\arabic{section}}
\counterwithin{figure}{section}
\counterwithin{table}{section}

\section{Colored-Noise Properties}

Noise samples drawn from a colored-noise process are correlated
between time steps when $\beta > 0$.

\begin{figure}[h]
  \begin{subfigure}{\linewidth}
    \includegraphics[width=.985\linewidth]{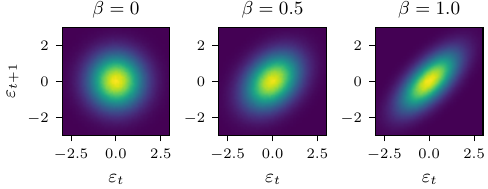}
    \caption{Probability of $\varepsilon_t$ vs. $\varepsilon_{t+1}$: subsequent noise samples are correlated when $\beta > 0$.
      (left) $\beta=0$, (middle) $\beta=0.5$ (right) $\beta=1.0$}
    \label{subfig:gaussian_correlated}
  \end{subfigure}
  \begin{subfigure}{\linewidth}
    \includegraphics[width=\linewidth]{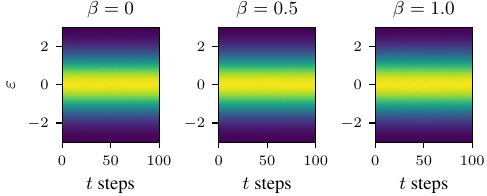}
    \caption{Averaging over sequences of $\varepsilon_t$ shows
      $\varepsilon_t$ is Gaussian distributed at each step. The noise
      color does not influence marginal distribution properties. }
    \label{subfig:noise-has-gaussian-limit}
  \end{subfigure}
  \caption{While colored noise shows increasing two-step correlation
    between $\varepsilon_t$ and $\varepsilon_{t+1}$ with increasing
    $\beta$, the marginal distribution of $\varepsilon_t$ at each
    time step remains Gaussian. }
    \label{fig:noise-has-gaussian-limit}
\end{figure}

\section{Best $\beta$ for $\nenv$}
\begin{table}[H]
  \centering
  \input{\ANALYSISPATH/best-noise-color-per-nenvs.tex}

\parbox{0.65\linewidth}{\caption{Noise color $\beta$ associated with
      the best average performance across environments. The listed
      performances were standardized to zero mean and unit variance
      for each environment.}
    \label{tbl:best-beta-for-nenv}
  }

\end{table}

\eclearpage
\section{Colored noise generation\label{sec:cngen}}

We build on the algorithm by \citet{timmerGeneratingPowerLaw1995} as
implemented by
\citet{patzeltFelixpatzeltColorednoise2019,eberhardPinkNoiseAll2023}:
we generate noise sequences $\tau_\varepsilon = \{\varepsilon_1 \ldots
\varepsilon_\tau\} $ of a pre-defined length $\tau$. See
\refalg{alg:colored_noise}.  In our experiments, we use sequences of
length $\tau=1000$, regenerating the whole sequence when all items are
consumed.

\begin{algorithm}
\caption{Generating Colored Noise using Inverse Fourier Transform.}
\label{alg:colored_noise}
\begin{algorithmic}[1]
  \Procedure{GenerateColoredNoise}{$N, \beta$}
  \State $L \gets \lfloor\sfrac{N}{2}\rfloor$
  \State $f \gets \{\frac{1}{N}, \frac{1}{N}, \ldots, \frac{i}{N}, \ldots, \frac{L}{N}\}$ \Comment{Frequencies of components $0\ldots L$}
  \State $s \gets \{\ldots, f_i^{\sfrac{-\beta}{2}}, \ldots\}$ \Comment{Calculate scales}
  \State $w_L \gets 
\begin{cases}
    s_L, & \text{if } L \text{ is odd}\\
    \sfrac{s_L}{2},              & \text{otherwise}
\end{cases}$
\State $w \gets \{s_1, \ldots s_{L-1}, w_L\}$
\State $\sigma \gets \frac{2}{N} \cdot \sqrt{\sum w_i^2}$
\State $\alpha = \{\ldots, \alpha_i, \ldots \}: \alpha_i \sim \mathcal{N}(0, s_i)$ \Comment{Real part}
\State $\beta = \{\ldots, \beta_i, \ldots \}: \beta_i \sim \mathcal{N}(0, s_i)$ \Comment{Imaginary part}
\State $\alpha_0 \sim \mathcal{N}(0, s_0 \cdot \sqrt{2})$
\State $\beta_0 \gets 0$
\State $\alpha_L \sim \begin{cases}
  \mathcal{N}(0, s_0 \cdot \sqrt{2}), & \text{if $odd$} \\
  \mathcal{N}(0, s_0), & \text{otherwise}
  \end{cases}
  $
\State $\beta_L \sim  \mathcal{N}(0, s_0) \cdot \begin{cases}
  0, & \text{if $odd$} \\
  1, & \text{otherwise}
  \end{cases}$
\State $\gamma \gets \{\ldots, \gamma_i, \ldots\}: \gamma_i = \alpha_i + \bm{i} \beta_i$
\State $\tau_\varepsilon = \mathcal{F}^{-1}[\gamma] \cdot \sfrac{1}{\sigma}$
\State \Return $\tau_\varepsilon$ \Comment{Return noise sequence of length $N$}
\EndProcedure
\end{algorithmic}
\end{algorithm}

\section{Benchmark Environments}
\begin{center}
  \centering
\parbox{0.65\linewidth}{\captionof{table}{Environment name used throughout the paper and exact spec-id used with \texttt{gym}}}
  \adjustbox{max width=\linewidth}{
    \input{\ANALYSISPATH/appendix_env_name_gym_spec.tex}

    }
\end{center}
\eclearpage

\section{Hyperparameters \& Experiment Details\label{sec:experiment_details}}
We use the PPO implementation by \citet{stable-baselines3}.
\begin{table}[H]
  \parbox{0.65\linewidth}{\caption{Hyperparameters used in our experiments and default values used by the PPO implementation \citep{stable-baselines3} when different.}}
  \centering
  \input{\ANALYSISPATH/hyperparameters.tex}

\end{table}
\section{Colored Noise $\beta=0$ matches Vanilla PPO}\label{sec:beta0vanilla}
\begin{figure}[H]
     \includegraphics[width=\linewidth]{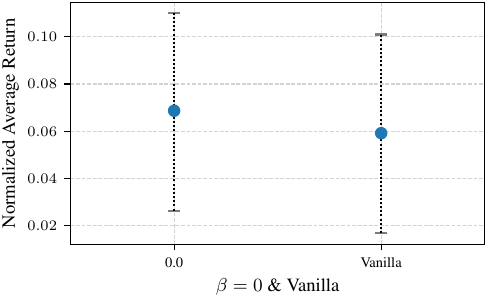}
    \caption{Gaussian noise as used in vanilla PPO corresponds to
      white ($\beta=0$) colored noise. Thus, the two algorithms are
      equivalent: Due to slightly different code paths minor
      deviations in performance estimates are expected, but the
      confidence intervals largely overlap and thus there is no
      significant difference between $\beta=0$ and the vanilla PPO
      implementation.}
    \label{fig:beta0vanilla}
\end{figure}

\eclearpage
\ifdefined\aaaiversion
  \changed{\section{Transfer to other methods}\label{sec:transfer_trpo}}

  \begin{figure}[H]
    \centering
    \begin{subfigure}[b]{0.31\linewidth}
      \centering
      \includegraphics[width=\linewidth]{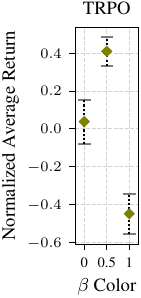}
      \caption{TRPO}
      \label{fig:image1}
    \end{subfigure}
    \hspace{0.01\linewidth}
    \begin{subfigure}[b]{0.31\linewidth}
      \centering
      \includegraphics[width=\linewidth]{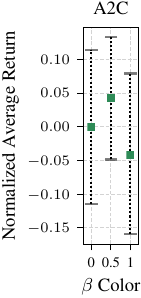}
      \caption{A2C}
      \label{fig:image2}
    \end{subfigure}
    \hspace{0.01\linewidth}
    \begin{subfigure}[b]{0.31\linewidth}
      \centering
      \includegraphics[width=\linewidth]{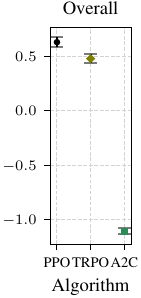}
      \caption{Overall}
      \label{fig:image2}
    \end{subfigure}
    \caption{\changed{Normalized averaged return across environments
        with $\nenv=4$. Performance is improved with $\beta=0.5$ also
        in the (a) TRPO method \citep{schulmanTrustRegionPolicy2015},
        (b) the A2C method \citep{mnihAsynchronousMethodsDeep2016},
        providing evidence for the transferability of our results. Due
        to its popularity the bulk of our investigation focuses on
        PPO. The effect is less pronounced in (b) A2C, but the overall
        comparison (c) shows that A2C was not able to learn as well as
        both PPO and TRPO.} }
    \label{fig:trpo_beta_better}
  \end{figure}

\fi

\ifdefined\aaaiversion
\changed{\section{Complex Tasks}\label{sec:complex_tasks}}
\urldef{\urlMujocoMaze}\url{https://github.com/kngwyu/mujoco-maze} \begin{figure}[H]
\includegraphics[width=\linewidth]{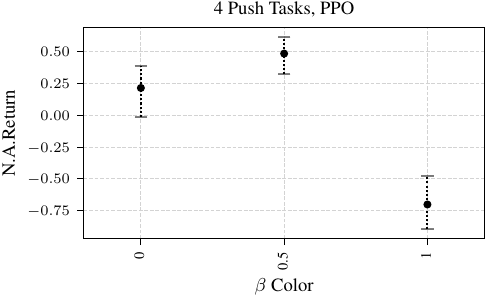}
\caption{\changed{The tendency to favor ${\beta\textrm{=}.5}$ also
    scales to complex tasks: the Push and PushMaze tasks of the mujoco
    maze\footnotemark
    benchmarks, with both the Point and Ant robots. In these tasks, the
    agent has to push one (Push) or multiple (PushMaze) obstacles out
    of the way before being able to reach a goal. The reward is a
    sparse goal reaching reward without a sub-task reward. In this
    setting, the agent is required to explore more globally, as there
    is little guiding information. Thus, we presume larger state-space
    coverage to help, and state-space coverage is increased by
    correlated action noise.
  }
  }
    \label{fig:complex_tasks}
\end{figure}
\footnotetext{\urlMujocoMaze}
\fi
\section{Final Returns \& Performance}

\begin{table}[H]
  \adjustbox{max width=\linewidth}{
    \input{./\ANALYSISPATH/appendix_final_returns.tex}

  }
\centering
  \parbox{\linewidth}{
  \caption{Return of the final policy (Final Return), \emph{mean/max}
    across noise colors and number of parallel collection
    environments. Return averaged across training (Performance),
    \emph{mean/max} across noise colors and number of parallel
    collection environments}}
\end{table}

\ifdefined\aaaiversion
\changed{\section{Effect of $\beta$ on learning curves}\label{sec:_learning_curves}}
\begin{figure}[H]
     \includegraphics[width=\linewidth]{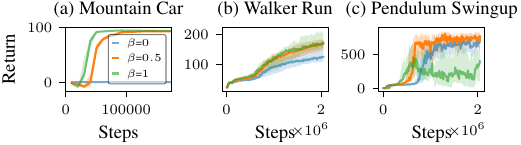}
     \caption{\changed{Examples of the effect of different $\beta$:
         (a) requires correlated noise for successful learning (x-axis
         is zoomed to highlight non-converged part), (b) benefits from
         more correlated noise, while in (c) more correlated noise
         destabilizes training, requiring a trade-off. Results for
         $\nenv=4$. Our experiments do not show collapse of superior
         policies: higher AUC thus means reaching better performance
         overall or reaching similar performance faster.  }}
     \label{fig:focused_impact_of_beta}
\end{figure}
\begin{figure}[H]
     \includegraphics[width=\linewidth]{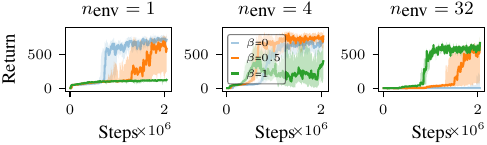}
     \caption{\changed{Examples of the effect of larger $\nenv$ on the
         efficacy of $\beta$ in the case of the Pendulum Swingup: the
         transition from $\beta=0$, to $\beta=0.5$ to $\beta=1$ with
         increasing $\nenv$ from $1$ over $4$ to $32$.  Learning
         curves for all environments are shown in
         \reffig{fig:focused_impact_of_nenv_and_beta_all}.  }}
     \label{fig:focused_impact_of_increasing_nenv}
\end{figure}
\begin{figure*}
     \includegraphics[width=\linewidth]{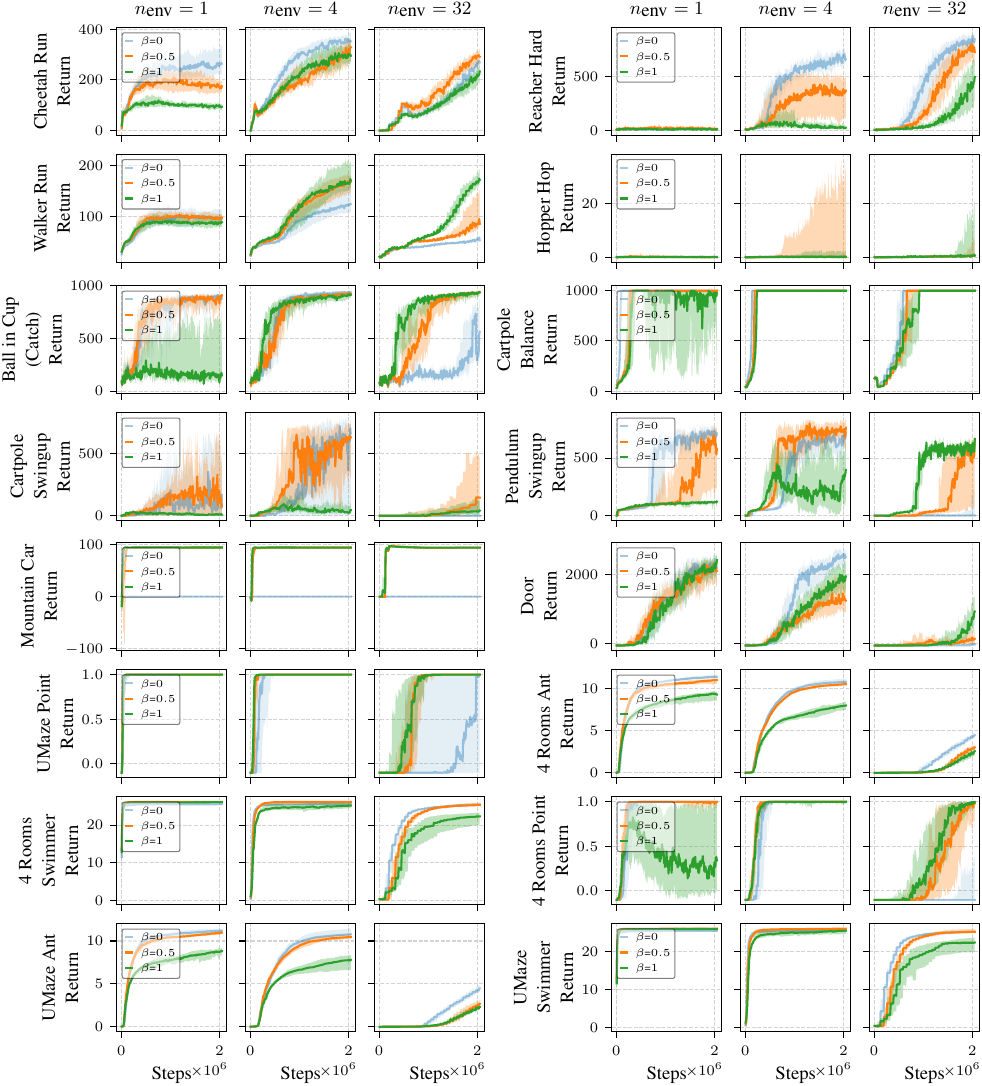}
     \caption{\changed{Learning curves for $\beta \in \{0, 0.5, 1\}$
         for $\nenv \in \{1,4,32\}$. With larger update sizes (i.e.,
         increasing $\nenv$), the relative efficacy of more correlated
         noise (i.e., $\beta=1$) tends to increase.  }}
     \label{fig:focused_impact_of_nenv_and_beta_all}
\end{figure*}
\fi

\section{Update Size Matters: $\nenv$ vs. $\nsteps$\label{sec:nenv_vs_nstep}}
\begin{figure}[H]
     \includegraphics[width=\linewidth]{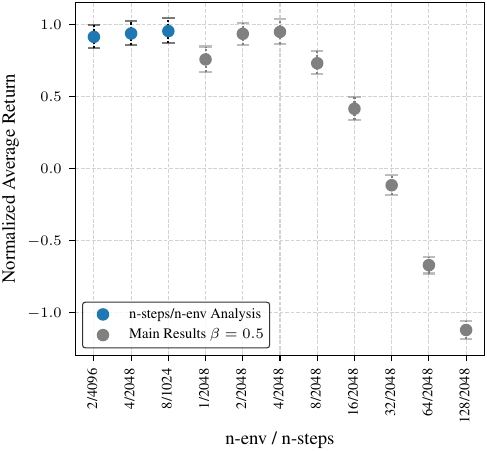}
    \caption{Analysis of the impact of $\nenv$ vs$ \nsteps$
        for $\beta=0.5$: When the total sample size is investigated
        ($\nenv \cdot \nsteps$), the influence of $\nenv$ and
        $\nsteps$ appears not significant (for $\nsteps~>~\textrm{episode length}$).}
    \label{fig:nenv_vs_nsteps}
\end{figure}

 In this work we found $\nenv = 4$ to perform best. This has
  to be viewed in combination with $\nsteps=2048$, which is the
  default value for the used implementation. Note that the
  environments used in our experiments have an episode length
  $<\nsteps$ and thus, the relevant quantity is the update size
  ($\nenv \cdot \nsteps$). This is confirmed by the experiment in
  \reffig{fig:nenv_vs_nsteps}. Which shows the main results (filtered
  by $\beta=0.5$) and additional experimental results ($\beta=0.5$)
  over various $\nenv \in \{2,4,8\}$ and matching changes in $\nsteps
  \in \{4096, 2048, 1024\}$ to keep the overall update size at
  $8192$. The results indicate that the change of $\nenv$
  vs. $\nsteps$ (when keeping the total update size constant), are not
  significant.

\section{$\nepochs$ vs $\beta$}
\begin{figure}[H]
     \includegraphics[width=\linewidth]{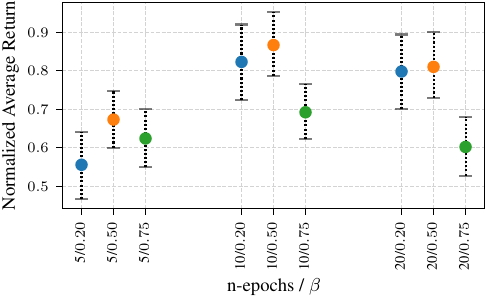}
     \caption{Analysis of the impact of $\nepochs$
         vs. $\beta$ on performance. $\beta \in \{0.2, 0.5, 0.75\}$,
         $\nenv=4$ and \mbox{$\nepochs \in \{5,10,20\}$}. The color
         preference does not appear to change significantly with variations in
         $\nepochs$.}
    \label{fig:beta_vs_nepochs}
\end{figure}

In this experiment we analyze the impact of varying $\nepochs \in
\{5,10,20\}$ where $10$ is the default used in the main
experiments. \reffig{fig:beta_vs_nepochs} shows that the color
preference for $\beta=0.5$ does not appear to change significantly with variation in
$\nepochs$. The experiments were performed with $\nenv=4$ on the same
environments and seeds as before. Halving $\nepochs$ reduces the
performance, while doubling the number of epochs does not appear to
have a significant impact on the color preference.

\ifdefined\aaaiversion
\section{SDE vs. Colored Noise}
\citet{raffinSmoothExplorationRobotic2021} propose to use generalized
State Dependent Exploration (gSDE), an extension over the method
proposed by \citet{ruckstiessStateDependentExplorationPolicy2008}. In
this experiment we compared gSDE exploration in PPO against colored
noise. We performed experiments on the same environments as before,
using the same number of seeds and variation over $\nenv$. The default
settings for gSDE defined in the implementation by
\citet{stable-baselines3} were used, i.e., gSDE was simply turned on by
setting \texttt{use\_sde} to \texttt{true}. \reffig{fig:sde_vs_colors}
shows the result.
\begin{figure}[H]
     \includegraphics[width=\linewidth]{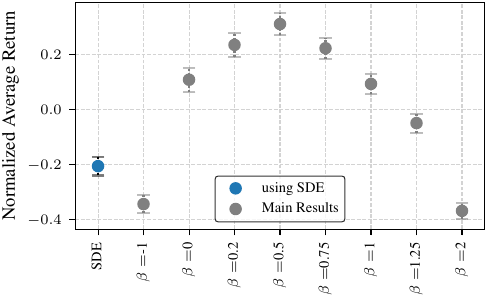}
     \caption{State Dependent Exploration vs. Colored Noise: SDE leads
       to more correlated actions and correspondingly yields result
       similar to more correlated noise (i.e., $\beta \in \{1.25,
       2\}$).}
    \label{fig:sde_vs_colors}
\end{figure}
\fi

\onecolumn
\section{Final Returns per $\beta$}
\begin{table}[H]
  \caption{Final Return per $\beta$ and $N_\textrm{envs}$. $\beta=0$
    is equivalent to the vanilla PPO implementation. Mean across seeds
    is reported.}
\end{table}
\vspace{-22pt}
       {
         \small
  \input{./\ANALYSISPATH/appendix_final_return_per_color.tex}

}
\twocolumn

 \end{document}

%% file: analysis/aaai/better_default_count_commands.tex
\newcommand{\betterdefaultnumexperiments}{16}
\newcommand{\isfavorablebetaWhite}[0]{8}
\newcommand{\isfavorablebetaIntermediate}[0]{11}
\newcommand{\isfavorablebetaPink}[0]{7}
\newcommand{\isoutperformedbetaWhite}[0]{8}
\newcommand{\isoutperformedbetaIntermediate}[0]{5}
\newcommand{\isoutperformedbetaPink}[0]{9}

\newcommand{\betterdefaultnumexperimentsW}{sixteen\xspace}
\newcommand{\isfavorablebetaWhiteW}[0]{eight\xspace}
\newcommand{\isfavorablebetaIntermediateW}[0]{eleven\xspace}
\newcommand{\isfavorablebetaPinkW}[0]{seven\xspace}
\newcommand{\isoutperformedbetaWhiteW}[0]{eight\xspace}
\newcommand{\isoutperformedbetaIntermediateW}[0]{five\xspace}
\newcommand{\isoutperformedbetaPinkW}[0]{nine\xspace}

%% file: analysis/aaai/used_env_names.tex
4 Rooms Ant, 4 Rooms Point, 4 Rooms Swimmer, Ball in Cup (Catch), Cartpole Balance, Cartpole Swingup, Cheetah Run, Door, Hopper Hop, Mountain Car, Pendulum Swingup, Reacher Hard, UMaze Ant, UMaze Point, UMaze Swimmer and Walker Run

%% file: analysis/aaai/used_noise_colors.tex
$-1$ (blue), $0$ (white), $0.2$, $0.5$, $0.75$, $1$ (pink), $1.25$, $2$ (red)

%% file: analysis/aaai/used_num_envs.tex
1, 2, 4, 8, 16, 32, 64, 128

%% file: analysis/aaai/total_number_of_experiments.tex
20480

%% file: analysis/aaai/best-noise-color-per-environment.tex
\begin{tabular}{@{}lcp{2.3em}p{1.3em}cp{1.3em}cp{1.3em}c@{}}
\toprule
Environment & $\beta^*$ & Noise Color & $\beta\newline 0.5$ & $p_{0.5}$ & $\beta\newline 0.0$ & $p_{0.0}$ & $\beta\newline 1.0$ & $p_{1.0}$ \\
\midrule
4 Rooms Swimmer & 0.00 & White & \cmark & 0.19 & \cmark & -- & \xmark & 0.01 \\
Cartpole Balance & 0.00 & White & \xmark & 0.04 & \cmark & -- & \xmark & 0.01 \\
Cheetah Run & 0.00 & White & \cmark & 0.14 & \cmark & -- & \xmark & 0.01 \\
Reacher Hard & 0.00 & White & \xmark & 0.01 & \cmark & -- & \xmark & 0.01 \\
UMaze Ant & 0.00 & White & \cmark & 0.24 & \cmark & -- & \xmark & 0.01 \\
UMaze Swimmer & 0.00 & White & \cmark & 0.16 & \cmark & -- & \xmark & 0.01 \\
4 Rooms Ant & 0.20 &  & \cmark & 0.21 & \cmark & 0.98 & \xmark & 0.01 \\
Door & 0.20 &  & \cmark & 0.06 & \cmark & 0.56 & \cmark & 0.25 \\
Cartpole Swingup & 0.50 &  & \cmark & -- & \xmark & 0.03 & \xmark & 0.01 \\
4 Rooms Point & 0.75 &  & \cmark & 0.97 & \xmark & 0.03 & \cmark & 0.51 \\
Ball in Cup (Catch) & 0.75 &  & \cmark & 0.90 & \xmark & 0.01 & \cmark & 0.96 \\
Hopper Hop & 0.75 &  & \cmark & 0.68 & \xmark & 0.01 & \cmark & 0.09 \\
Pendulum Swingup & 0.75 &  & \cmark & 0.92 & \xmark & 0.04 & \xmark & 0.03 \\
Walker Run & 1.00 & Pink & \xmark & 0.01 & \xmark & 0.01 & \cmark & -- \\
Mountain Car & 1.25 &  & \xmark & 0.01 & \xmark & 0.01 & \cmark & 0.45 \\
UMaze Point & 2.00 & Red & \xmark & 0.01 & \xmark & 0.01 & \cmark & 0.06 \\
\bottomrule
\end{tabular}

%% file: analysis/aaai/best-noise-color-per-nenvs.tex
\begin{tabular}{@{}rrr@{}}
\toprule
$\nenv$ & $\beta$ & Average Performance \\
\midrule
1 & 0.2 & 0.81 \\
2 & 0.5 & 0.99 \\
4 & 0.5 & 1.01 \\
8 & 0.75 & 0.79 \\
16 & 0.75 & 0.51 \\
32 & 0.75 & -0.04 \\
64 & 0.5 & -0.64 \\
128 & 2 & -1 \\
\bottomrule
\end{tabular}

%% file: analysis/aaai/appendix_env_name_gym_spec.tex
\begin{tabular}{@{}lr@{}}
\toprule
Environment & Gym Spec. \\
\midrule
Mountain Car & MountainCarContinuous-v0 \\
Ball in Cup (Catch) & dm2gym.envs:Ball\_in\_cupCatch-v0 \\
Cartpole Balance & dm2gym.envs:CartpoleBalance\_sparse-v0 \\
Cartpole Swingup & dm2gym.envs:CartpoleSwingup\_sparse-v0 \\
Cheetah Run & dm2gym.envs:CheetahRun-v0 \\
Hopper Hop & dm2gym.envs:HopperHop-v0 \\
Pendulum Swingup & dm2gym.envs:PendulumSwingup-v0 \\
Reacher Hard & dm2gym.envs:ReacherHard-v0 \\
Walker Run & dm2gym.envs:WalkerRun-v0 \\
Door & d4rl.hand\_manipulation\_suite:door-v0 \\
4 Rooms Swimmer & mujoco\_maze:Swimmer4Rooms-v1 \\
4 Rooms Point & mujoco\_maze:Point4Rooms-v1 \\
4 Rooms Ant & mujoco\_maze:Ant4Rooms-v1 \\
UMaze Ant & mujoco\_maze:AntUMaze-v1 \\
UMaze Swimmer & mujoco\_maze:SwimmerUMaze-v1 \\
UMaze Point & mujoco\_maze:PointUMaze-v1 \\
\bottomrule
\end{tabular}

%% file: analysis/aaai/hyperparameters.tex
\begin{tabular}{@{}lrr@{}}
\toprule
Param. & Value & Default \\
\midrule
lr & 0.00025 & 0.0003 \\
n-steps & 2048 &  \\
batch-size & 128 & 64 \\
n-epochs & 10 &  \\
gamma & 0.99 &  \\
gae-lambda & 0.95 &  \\
clip-range & 0.2 &  \\
normalize-advantage & True &  \\
ent-coef & 0 &  \\
vf-coef & 0.5 &  \\
max-grad-norm & 0.5 &  \\
use-sde & False &  \\
sde-sample-freq & -1 &  \\
stats-window-size & 100 &  \\
\bottomrule
\end{tabular}

%% file: analysis/aaai/appendix_final_returns.tex
\begin{tabular}{@{}lrrrr@{}}
\toprule
 & \multicolumn{2}{r}{Final Return} & \multicolumn{2}{r}{Performance} \\
 & mean & max & mean & max \\
env &  &  &  &  \\
\midrule
4 Rooms Ant & 5.50 & 12.16 & 3.72 & 10.50 \\
4 Rooms Point & 0.58 & 1.00 & 0.40 & 0.97 \\
4 Rooms Swimmer & 22.91 & 26.22 & 19.49 & 26.08 \\
Ball in Cup (Catch) & 685.85 & 966.92 & 500.22 & 869.97 \\
Cartpole Balance & 742.18 & 1000.00 & 608.22 & 976.19 \\
Cartpole Swingup & 126.55 & 843.10 & 75.24 & 627.93 \\
Cheetah Run & 222.70 & 558.55 & 139.34 & 351.39 \\
Door & 921.96 & 3128.73 & 372.29 & 1937.42 \\
Hopper Hop & 5.44 & 77.86 & 2.39 & 34.25 \\
Mountain Car & 66.92 & 98.33 & 61.88 & 93.53 \\
Pendulum Swingup & 338.21 & 885.04 & 205.43 & 678.91 \\
Reacher Hard & 266.71 & 935.94 & 139.72 & 613.31 \\
UMaze Ant & 5.49 & 12.54 & 3.70 & 10.78 \\
UMaze Point & 0.85 & 1.00 & 0.67 & 0.99 \\
UMaze Swimmer & 22.83 & 26.22 & 19.47 & 26.08 \\
Walker Run & 105.10 & 473.11 & 69.66 & 194.53 \\
\bottomrule
\end{tabular}

%% file: analysis/aaai/appendix_final_return_per_color.tex
\begin{longtable}{@{}lrrrrrrrrrr@{}}
\toprule
 & $\beta$ & -1.00 & 0.00 & 0.20 & 0.50 & 0.75 & 1.00 & 1.25 & 2.00 & Vanilla \\
env & n-envs &  &  &  &  &  &  &  &  &  \\
\midrule
\endfirsthead
\toprule
 & $\beta$ & -1.00 & 0.00 & 0.20 & 0.50 & 0.75 & 1.00 & 1.25 & 2.00 & Vanilla \\
env & n-envs &  &  &  &  &  &  &  &  &  \\
\midrule
\endhead
\midrule
\multicolumn{11}{r}{Continued on next page} \\
\midrule
\endfoot
\bottomrule
\endlastfoot
\multirow[t]{8}{*}{4 Rooms Ant} & 1 & 8.8 & 11.4 & 11.3 & 10.9 & 9.9 & 9.1 & 7.6 & 7.5 & 11.3 \\
 & 2 & 8.6 & 11.1 & 11.3 & 10.8 & 9.9 & 8.6 & 7.2 & 5.2 & 11.2 \\
 & 4 & 8.1 & 10.9 & 11.0 & 10.6 & 9.5 & 7.9 & 6.1 & 4.4 & 10.9 \\
 & 8 & 6.5 & 10.6 & 10.7 & 9.3 & 8.3 & 7.0 & 5.7 & 3.9 & 10.4 \\
 & 16 & 4.4 & 8.2 & 8.5 & 6.2 & 5.5 & 4.9 & 4.7 & 4.0 & 8.6 \\
 & 32 & 2.3 & 4.5 & 4.0 & 2.9 & 2.8 & 2.4 & 2.3 & 2.9 & 4.5 \\
 & 64 & 1.1 & 0.7 & 0.1 & 0.0 & 0.1 & 0.0 & 0.1 & 0.3 & 0.7 \\
 & 128 & 0.2 & -0.0 & -0.0 & -0.0 & -0.0 & -0.1 & -0.1 & -0.0 & -0.0 \\
\cline{1-11}
\multirow[t]{8}{*}{4 Rooms Point} & 1 & 0.9 & 0.9 & 0.8 & 0.9 & 0.5 & 0.4 & 0.0 & 0.1 & 0.9 \\
 & 2 & 0.9 & 0.9 & 0.9 & 0.9 & 0.9 & 0.8 & 0.5 & 0.1 & 0.9 \\
 & 4 & 1.0 & 0.9 & 0.8 & 0.9 & 0.9 & 0.9 & 0.8 & 0.4 & 0.9 \\
 & 8 & 0.8 & 0.9 & 0.9 & 0.9 & 0.9 & 0.9 & 0.9 & 0.8 & 0.8 \\
 & 16 & 0.7 & 0.8 & 0.8 & 0.9 & 1.0 & 0.9 & 0.9 & 0.9 & 0.7 \\
 & 32 & 0.2 & 0.2 & 0.6 & 0.8 & 0.9 & 0.9 & 0.9 & 0.8 & 0.3 \\
 & 64 & -0.0 & 0.0 & 0.0 & 0.0 & 0.3 & 0.3 & 0.3 & 0.3 & -0.0 \\
 & 128 & -0.1 & -0.1 & -0.1 & -0.1 & -0.1 & -0.1 & 0.0 & -0.0 & -0.1 \\
\cline{1-11}
\multirow[t]{8}{*}{4 Rooms Swimmer} & 1 & 24.8 & 25.6 & 25.8 & 26.0 & 26.0 & 26.0 & 25.5 & 24.5 & 25.5 \\
 & 2 & 24.7 & 25.6 & 25.8 & 26.0 & 25.8 & 25.7 & 25.5 & 23.0 & 25.6 \\
 & 4 & 24.4 & 25.5 & 25.8 & 25.9 & 25.7 & 25.1 & 24.5 & 22.7 & 25.5 \\
 & 8 & 24.3 & 25.5 & 25.7 & 25.8 & 25.3 & 24.5 & 23.0 & 20.6 & 25.5 \\
 & 16 & 23.7 & 25.4 & 25.5 & 25.5 & 25.2 & 23.5 & 22.0 & 18.1 & 25.4 \\
 & 32 & 23.0 & 25.3 & 25.7 & 25.4 & 23.5 & 21.4 & 21.3 & 18.7 & 25.3 \\
 & 64 & 21.2 & 24.6 & 24.7 & 24.1 & 19.2 & 18.8 & 18.8 & 17.3 & 24.6 \\
 & 128 & 14.3 & 21.1 & 19.9 & 17.4 & 13.2 & 12.2 & 12.2 & 13.3 & 20.9 \\
\cline{1-11}
\multirow[t]{8}{*}{Ball in Cup (Catch)} & 1 & 701.4 & 833.4 & 843.5 & 811.4 & 721.8 & 308.4 & 356.1 & 290.8 & 811.4 \\
 & 2 & 826.3 & 816.7 & 900.4 & 863.6 & 798.3 & 751.1 & 376.5 & 386.4 & 888.5 \\
 & 4 & 887.7 & 936.8 & 929.1 & 922.8 & 829.2 & 861.8 & 584.3 & 393.3 & 926.6 \\
 & 8 & 886.6 & 933.2 & 921.9 & 930.4 & 887.7 & 766.8 & 793.4 & 403.4 & 896.8 \\
 & 16 & 799.0 & 881.0 & 932.2 & 938.6 & 906.4 & 913.8 & 922.3 & 385.8 & 884.8 \\
 & 32 & 570.2 & 575.2 & 829.0 & 907.4 & 929.0 & 891.9 & 832.6 & 532.3 & 488.4 \\
 & 64 & 451.5 & 146.5 & 349.4 & 721.5 & 688.3 & 753.8 & 855.9 & 664.3 & 140.0 \\
 & 128 & 225.7 & 126.0 & 133.5 & 218.7 & 306.0 & 574.7 & 733.8 & 743.2 & 127.4 \\
\cline{1-11}
\multirow[t]{8}{*}{Cartpole Balance} & 1 & 820.6 & 981.2 & 1000.0 & 1000.0 & 810.1 & 691.5 & 171.1 & 51.0 & 973.4 \\
 & 2 & 902.7 & 992.5 & 1000.0 & 1000.0 & 998.2 & 983.6 & 637.7 & 59.4 & 997.2 \\
 & 4 & 892.3 & 1000.0 & 1000.0 & 1000.0 & 999.0 & 988.1 & 799.1 & 49.0 & 998.4 \\
 & 8 & 762.2 & 996.7 & 1000.0 & 986.1 & 1000.0 & 999.4 & 743.8 & 89.0 & 1000.0 \\
 & 16 & 221.3 & 1000.0 & 1000.0 & 990.8 & 1000.0 & 1000.0 & 723.8 & 35.4 & 1000.0 \\
 & 32 & 153.9 & 990.7 & 1000.0 & 1000.0 & 1000.0 & 1000.0 & 901.3 & 96.7 & 1000.0 \\
 & 64 & 115.0 & 967.3 & 1000.0 & 1000.0 & 1000.0 & 920.8 & 860.2 & 418.3 & 1000.0 \\
 & 128 & 99.5 & 865.7 & 771.4 & 569.9 & 525.6 & 319.4 & 322.7 & 225.3 & 711.4 \\
\cline{1-11}
\multirow[t]{8}{*}{Cartpole Swingup} & 1 & 45.0 & 275.8 & 347.6 & 221.1 & 29.6 & 15.2 & 8.6 & 3.3 & 239.5 \\
 & 2 & 55.6 & 404.4 & 534.2 & 337.8 & 146.9 & 25.1 & 26.3 & 10.2 & 268.8 \\
 & 4 & 27.4 & 469.8 & 488.0 & 501.7 & 113.6 & 70.0 & 24.4 & 7.2 & 461.3 \\
 & 8 & 27.5 & 407.5 & 393.8 & 365.3 & 118.8 & 29.0 & 19.7 & 13.9 & 430.5 \\
 & 16 & 39.0 & 372.1 & 328.5 & 320.9 & 159.9 & 55.9 & 29.8 & 19.8 & 264.0 \\
 & 32 & 11.9 & 119.1 & 235.0 & 264.7 & 149.5 & 71.9 & 38.6 & 10.3 & 191.4 \\
 & 64 & 4.5 & 19.8 & 23.1 & 72.0 & 35.8 & 46.7 & 28.2 & 15.7 & 16.6 \\
 & 128 & 0.7 & 6.1 & 7.0 & 5.4 & 9.0 & 14.0 & 5.8 & 14.0 & 4.1 \\
\cline{1-11}
\multirow[t]{8}{*}{Cheetah Run} & 1 & 220.8 & 279.5 & 242.9 & 184.6 & 145.6 & 95.3 & 99.9 & 124.6 & 240.0 \\
 & 2 & 273.5 & 327.2 & 316.8 & 290.0 & 206.2 & 170.3 & 152.3 & 162.2 & 331.4 \\
 & 4 & 311.6 & 357.2 & 357.3 & 311.6 & 326.8 & 299.0 & 239.6 & 217.2 & 337.7 \\
 & 8 & 314.2 & 368.1 & 330.3 & 320.7 & 348.4 & 350.9 & 320.1 & 223.5 & 362.6 \\
 & 16 & 289.9 & 334.3 & 328.6 & 354.5 & 338.4 & 323.7 & 297.5 & 200.5 & 358.8 \\
 & 32 & 204.4 & 276.2 & 294.7 & 296.4 & 254.2 & 230.3 & 166.5 & 100.0 & 242.8 \\
 & 64 & 147.8 & 118.5 & 178.6 & 182.2 & 166.7 & 93.1 & 70.3 & 39.9 & 131.7 \\
 & 128 & 84.2 & 123.5 & 128.9 & 104.9 & 87.1 & 64.8 & 45.9 & 37.8 & 127.2 \\
\cline{1-11}
\multirow[t]{8}{*}{Door} & 1 & -36.1 & 2061.2 & 2402.2 & 1978.1 & 2166.1 & 2239.9 & 1842.5 & 283.1 & 1968.0 \\
 & 2 & -15.3 & 2416.0 & 2142.0 & 1614.9 & 1765.8 & 1912.8 & 1991.2 & 284.1 & 2453.9 \\
 & 4 & 0.8 & 2451.7 & 2297.0 & 1427.0 & 1425.2 & 1884.0 & 1975.6 & 1046.5 & 2387.3 \\
 & 8 & -41.0 & 1928.3 & 1970.5 & 1423.9 & 1410.1 & 1696.5 & 1644.4 & 992.3 & 2285.7 \\
 & 16 & -49.4 & 588.5 & 1165.7 & 1333.0 & 1173.8 & 1300.2 & 1402.2 & 947.0 & 755.9 \\
 & 32 & -50.5 & 46.3 & 96.8 & 248.7 & 865.4 & 937.7 & 660.6 & -5.7 & 14.0 \\
 & 64 & -51.3 & -36.3 & 7.1 & 112.6 & 56.5 & 94.8 & 26.1 & -52.1 & -37.2 \\
 & 128 & -51.4 & -50.3 & -39.0 & -48.0 & -48.8 & -51.6 & -52.3 & -52.8 & -47.3 \\
\cline{1-11}
\multirow[t]{8}{*}{Hopper Hop} & 1 & 3.7 & 2.4 & 1.6 & 4.3 & 3.4 & 0.4 & 1.0 & 2.0 & 2.4 \\
 & 2 & 3.7 & 1.2 & 2.5 & 3.0 & 4.0 & 5.0 & 2.3 & 2.5 & 1.7 \\
 & 4 & 9.5 & 5.2 & 3.3 & 14.1 & 7.9 & 6.5 & 9.2 & 9.0 & 6.5 \\
 & 8 & 4.9 & 5.9 & 12.1 & 16.2 & 28.7 & 10.3 & 9.9 & 5.1 & 7.8 \\
 & 16 & 7.6 & 4.7 & 9.8 & 23.2 & 21.4 & 14.7 & 13.6 & 10.6 & 2.8 \\
 & 32 & 4.0 & 3.5 & 3.1 & 5.1 & 5.9 & 7.5 & 5.5 & 3.4 & 0.4 \\
 & 64 & 0.3 & 0.3 & 0.3 & 0.3 & 0.3 & 0.4 & 0.6 & 0.2 & 0.3 \\
 & 128 & 0.1 & 0.2 & 0.2 & 0.2 & 0.2 & 0.2 & 0.2 & 0.2 & 0.2 \\
\cline{1-11}
\multirow[t]{8}{*}{Mountain Car} & 1 & -0.0 & -0.0 & 28.0 & 93.9 & 93.9 & 93.9 & 93.8 & 93.8 & -0.0 \\
 & 2 & -0.0 & -0.0 & 32.4 & 93.9 & 93.9 & 93.8 & 93.9 & 93.9 & -0.0 \\
 & 4 & -0.0 & -0.0 & 23.4 & 93.9 & 93.8 & 93.8 & 93.8 & 93.8 & -0.0 \\
 & 8 & -0.0 & -0.0 & 84.6 & 93.9 & 93.8 & 93.8 & 93.8 & 93.7 & -0.0 \\
 & 16 & -0.0 & -0.0 & 79.9 & 93.9 & 93.8 & 93.8 & 93.8 & 93.6 & 4.7 \\
 & 32 & -0.0 & -0.0 & 93.8 & 93.8 & 93.9 & 93.8 & 93.8 & 93.4 & -0.0 \\
 & 64 & -0.0 & -0.0 & 95.7 & 93.2 & 93.5 & 93.5 & 93.5 & 92.5 & -0.0 \\
 & 128 & -0.0 & -0.0 & 96.7 & 95.0 & 94.3 & 93.6 & 92.7 & 91.9 & -0.0 \\
\cline{1-11}
\multirow[t]{8}{*}{Pendulum Swingup} & 1 & 501.9 & 714.5 & 648.0 & 521.1 & 328.5 & 153.2 & 106.3 & 58.5 & 697.2 \\
 & 2 & 401.4 & 673.6 & 675.2 & 646.3 & 362.6 & 168.9 & 165.5 & 101.1 & 737.0 \\
 & 4 & 420.3 & 677.5 & 766.9 & 694.6 & 625.0 & 403.1 & 270.7 & 153.1 & 733.1 \\
 & 8 & 333.9 & 655.1 & 694.4 & 675.7 & 554.9 & 210.9 & 152.8 & 56.5 & 642.5 \\
 & 16 & 149.9 & 370.7 & 432.6 & 640.7 & 646.9 & 379.6 & 207.3 & 29.7 & 381.5 \\
 & 32 & 32.6 & 192.3 & 221.6 & 441.9 & 582.9 & 602.7 & 413.1 & 304.5 & 135.3 \\
 & 64 & 9.8 & 4.4 & 10.4 & 71.5 & 367.8 & 494.0 & 526.3 & 523.5 & 7.6 \\
 & 128 & 2.0 & 1.2 & 1.1 & 2.2 & 15.1 & 69.5 & 126.2 & 199.0 & 1.1 \\
\cline{1-11}
\multirow[t]{8}{*}{Reacher Hard} & 1 & 10.9 & 22.6 & 30.5 & 19.0 & 14.6 & 17.5 & 13.4 & 12.1 & 66.4 \\
 & 2 & 55.1 & 426.0 & 291.9 & 186.0 & 43.4 & 18.9 & 20.9 & 16.3 & 350.2 \\
 & 4 & 133.8 & 663.8 & 513.8 & 332.5 & 188.5 & 45.5 & 37.2 & 19.0 & 559.5 \\
 & 8 & 292.0 & 696.4 & 516.2 & 533.3 & 310.8 & 133.2 & 76.2 & 60.7 & 692.0 \\
 & 16 & 322.5 & 776.5 & 718.1 & 654.4 & 499.5 & 422.7 & 308.4 & 169.7 & 742.9 \\
 & 32 & 341.5 & 833.6 & 791.8 & 712.1 & 600.5 & 447.4 & 264.7 & 188.9 & 665.7 \\
 & 64 & 207.2 & 551.8 & 597.6 & 546.7 & 283.2 & 132.1 & 105.3 & 150.8 & 499.4 \\
 & 128 & 55.4 & 208.8 & 204.4 & 78.4 & 35.7 & 30.7 & 20.7 & 56.2 & 237.1 \\
\cline{1-11}
\multirow[t]{8}{*}{UMaze Ant} & 1 & 8.9 & 11.2 & 11.2 & 10.9 & 10.1 & 8.6 & 7.4 & 7.4 & 11.1 \\
 & 2 & 8.3 & 11.1 & 11.0 & 10.8 & 10.0 & 8.6 & 6.8 & 5.8 & 11.1 \\
 & 4 & 7.9 & 11.0 & 10.9 & 10.5 & 9.5 & 7.6 & 6.1 & 4.0 & 11.0 \\
 & 8 & 6.5 & 10.5 & 10.7 & 9.5 & 8.4 & 7.0 & 6.2 & 3.7 & 10.4 \\
 & 16 & 4.5 & 8.5 & 8.5 & 6.6 & 5.8 & 5.0 & 4.6 & 4.1 & 8.3 \\
 & 32 & 2.4 & 4.4 & 3.8 & 2.7 & 2.7 & 2.3 & 2.3 & 2.8 & 4.5 \\
 & 64 & 1.2 & 0.6 & 0.1 & 0.0 & 0.1 & 0.0 & 0.1 & 0.3 & 0.6 \\
 & 128 & 0.1 & -0.0 & -0.0 & -0.0 & -0.0 & -0.1 & -0.1 & -0.1 & -0.0 \\
\cline{1-11}
\multirow[t]{8}{*}{UMaze Point} & 1 & 1.0 & 1.0 & 1.0 & 1.0 & 1.0 & 1.0 & 1.0 & 1.0 & 1.0 \\
 & 2 & 1.0 & 1.0 & 1.0 & 1.0 & 1.0 & 1.0 & 1.0 & 1.0 & 1.0 \\
 & 4 & 1.0 & 1.0 & 1.0 & 1.0 & 1.0 & 1.0 & 1.0 & 1.0 & 0.9 \\
 & 8 & 1.0 & 1.0 & 0.9 & 1.0 & 1.0 & 1.0 & 1.0 & 1.0 & 0.9 \\
 & 16 & 0.9 & 0.7 & 1.0 & 1.0 & 1.0 & 1.0 & 1.0 & 1.0 & 0.9 \\
 & 32 & 0.7 & 0.5 & 0.9 & 1.0 & 1.0 & 1.0 & 1.0 & 1.0 & 0.5 \\
 & 64 & 0.3 & 0.3 & 0.5 & 0.9 & 1.0 & 1.0 & 1.0 & 1.0 & 0.3 \\
 & 128 & 0.1 & 0.2 & 0.2 & 0.2 & 0.3 & 0.3 & 0.4 & 0.6 & 0.2 \\
\cline{1-11}
\multirow[t]{8}{*}{UMaze Swimmer} & 1 & 24.8 & 25.7 & 25.8 & 26.0 & 25.9 & 25.9 & 25.5 & 24.8 & 25.5 \\
 & 2 & 24.6 & 25.6 & 25.8 & 25.9 & 25.9 & 25.7 & 25.0 & 21.9 & 25.6 \\
 & 4 & 24.6 & 25.5 & 25.9 & 25.8 & 25.7 & 25.3 & 24.7 & 22.2 & 25.6 \\
 & 8 & 23.8 & 25.5 & 25.7 & 25.7 & 25.6 & 24.6 & 22.4 & 19.2 & 25.5 \\
 & 16 & 23.8 & 25.5 & 25.6 & 25.6 & 25.1 & 23.5 & 21.7 & 19.2 & 25.4 \\
 & 32 & 22.8 & 25.3 & 25.6 & 25.4 & 23.5 & 21.5 & 21.1 & 17.9 & 25.3 \\
 & 64 & 20.9 & 24.6 & 24.8 & 24.2 & 19.2 & 18.9 & 19.3 & 16.6 & 24.6 \\
 & 128 & 13.3 & 21.2 & 19.8 & 17.5 & 12.9 & 12.1 & 12.6 & 12.9 & 21.0 \\
\cline{1-11}
\multirow[t]{8}{*}{Walker Run} & 1 & 39.2 & 97.6 & 100.5 & 96.5 & 93.3 & 89.1 & 82.8 & 84.5 & 103.8 \\
 & 2 & 48.7 & 126.4 & 122.4 & 141.8 & 127.0 & 130.3 & 102.8 & 82.2 & 122.2 \\
 & 4 & 75.9 & 131.1 & 141.5 & 173.7 & 169.3 & 178.2 & 145.7 & 87.7 & 117.4 \\
 & 8 & 84.8 & 95.6 & 116.4 & 167.9 & 222.2 & 261.4 & 191.8 & 89.4 & 111.2 \\
 & 16 & 70.0 & 68.2 & 71.5 & 126.5 & 251.2 & 269.0 & 196.2 & 76.1 & 66.9 \\
 & 32 & 59.4 & 55.0 & 60.8 & 117.8 & 188.2 & 181.6 & 140.1 & 67.2 & 57.5 \\
 & 64 & 55.7 & 46.2 & 51.4 & 61.5 & 70.9 & 73.3 & 68.8 & 55.3 & 47.3 \\
 & 128 & 41.6 & 39.5 & 41.4 & 48.7 & 46.8 & 42.7 & 43.1 & 43.2 & 39.3 \\
\cline{1-11}
\end{longtable}